\pdfoutput=1

\documentclass[11pt]{article}

\usepackage[]{EMNLP2023}

\usepackage{times}
\usepackage{latexsym}

\usepackage[T1]{fontenc}

\usepackage[utf8]{inputenc}

\usepackage{microtype}

\usepackage{inconsolata}

\usepackage{graphicx}
\usepackage{tabularx}
\usepackage{multirow}

\title{Think Before You Speak: Cultivating Communication Skills of \\ Large Language Models via Inner Monologue}

\author{
Junkai Zhou$^{1,2}$,
Liang Pang$^{1}$\thanks{\ \ Corresponding authors},
Huawei Shen$^{1,2}$,
Xueqi Cheng$^{1,2}$\\
$^{1}$CAS Key Laboratory of AI Security, \\
Institute of Computing Technology, Chinese Academy of Sciences, Beijing, China \\
 $^{2}$University of Chinese Academy of Sciences, Beijing, China \\
{\tt\ \{zhoujunkai20z,pangliang,shenhuawei,cxq\}@ict.ac.cn}}

\begin{document}
\maketitle
\begin{abstract}
The emergence of large language models (LLMs) further improves the capabilities of open-domain dialogue systems and can generate fluent, coherent, and diverse responses.
However, LLMs still lack a crucial ability: communication skills. This limitation renders them more like information seeking tools rather than anthropomorphic chatbots.
Communication skills, such as topic transition, proactively asking questions, concept guidance, empathy, and summarising often should be taken into consideration, to make LLMs more anthropomorphic and proactive during the conversation, thereby increasing the interest of users and attracting them to chat for longer.
However, enabling these communication skills in black-box LLMs remains a key challenge because they do not have the same utterance formation mode as real people: think before speaking.
Inspired by linguistics and cognitive science, we empower LLMs with communication skills through inner monologues.
To evaluate various communication skills, we construct a benchmark named Cskills, which can also more comprehensively evaluate the dialogue generation ability of the model.
Experimental results show that the proposed CSIM strategy improves the backbone models and outperforms the baselines.
\end{abstract}

\section{Introduction}

\begin{figure}
    \centering 
    \includegraphics[width=8.05cm]{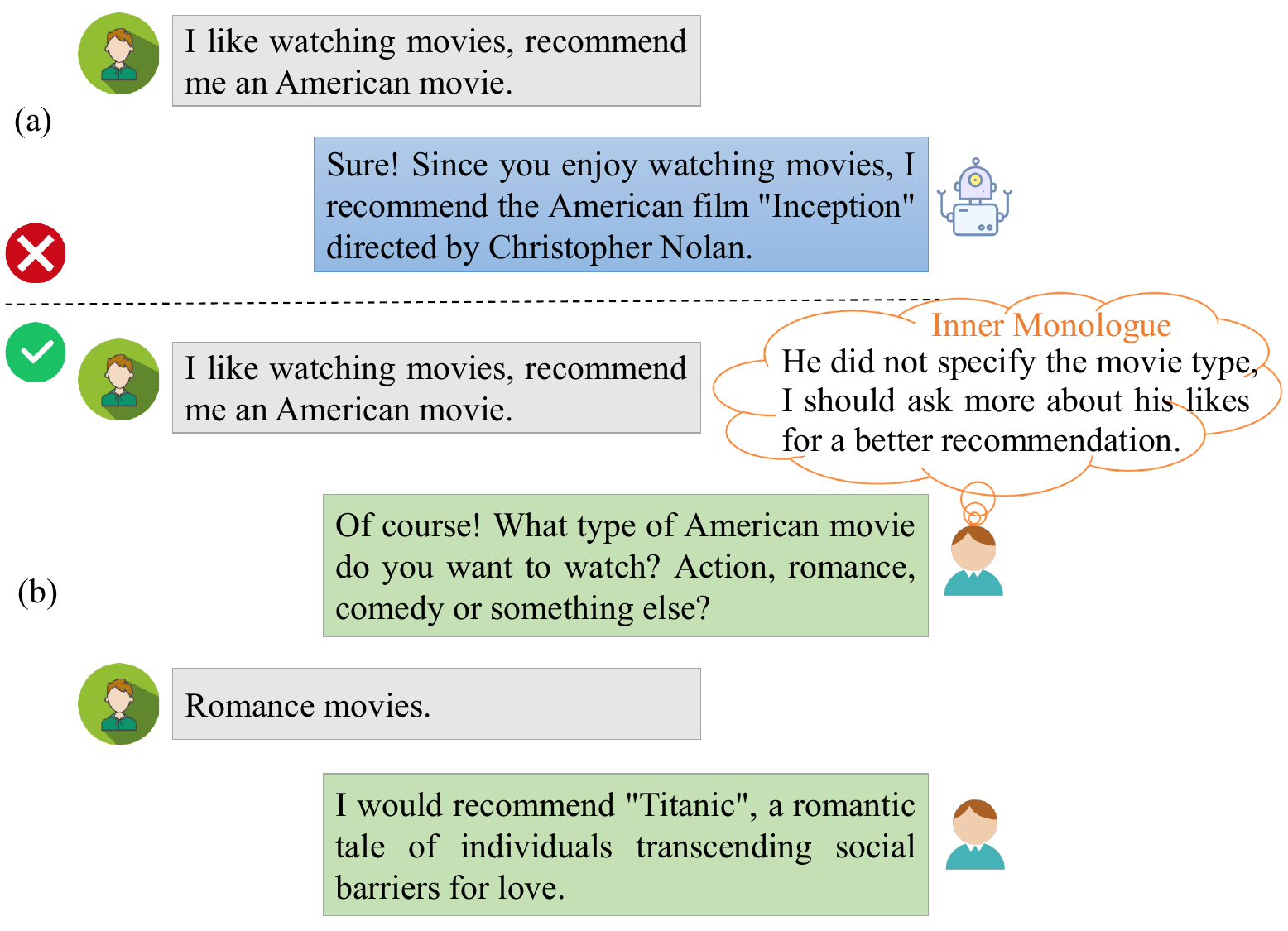}
    \caption{
    When asked to recommend: (a) ChatGPT directly recommends without asking the detailed needs of users, which may lead to failure to satisfy users; (b) people proactively ask questions to further understand the needs of users before making recommendations.
    }
    \label{fig:fig1}
\end{figure}

Open-domain dialogue systems need to generate fluent, coherent, and diverse responses based on history utterances.
The emergence of large language models \citep{chowdhery2022palm, openai2022chatgpt, touvron2023llama} further enhances the capabilities of dialogue generation systems and can meet the above requirements.
However, LLMs are more like an information seeking tool than a chatbot like a real person.
Such a dialogue system may make users lose interest in chatting and terminate the conversation.
The reason is that LLMs still lack an important conversational ability: communication skills. As shown in Figure~\ref{fig:fig1}, LLM makes recommendations without a thorough comprehension of the preferences of the user regarding movie genres. This lack of detailed understanding may result in inaccurate recommendation outcomes.
People use proactively asking questions in communication skills to further understand the needs of the user, thereby making better recommendations.

In linguistics, communication skills are used to enhance the interactive experience during the conversation and to establish effective communication \citep{dornyei1995teachability, grover2005shaping, barker2010improve}. 
The five common communication skills are topic transition, proactively asking questions, concept guidance, empathy, and summarising often.
Each communication skill is applicable to different conversational situations and plays a different role during the conversation.
By using topic transition \citep{dornyei1995teachability,riou2015methodology}, we can avoid unfamiliar concepts and transition to familiar ones, leading to better conversations. 
Proactively asking questions \citep{grover2005shaping} can help us clarify ambiguous information and make appropriate responses based on it.
Concept guidance \citep{zou2021thinking} can strengthen the connection of concepts in a conversation and increase the proactivity of the dialogue. 
Empathy \citep{rizzolatti2008mirrors} can produce more personal and informative responses, increasing the interest of the speaker in chatting.
Summarising often \citep{barker2010improve} allows speakers to confirm whether a consensus has been reached on the previous information, reducing the occurrence of misunderstandings.


Introducing communication skills to LLMs is not easy because they do not have the same utterance formation mode as real people: think before speaking. That is, LLMs do not have the same thinking process as real people before generating responses. LLMs are black-boxes makes understanding their decision-making process more challenging.
Existing works in psychology and cognitive science indicate that humans think before speaking when they have a conversation \citep{hulme1999think, khawaja2008think, neustein2012think}. \citet{LI2020102468} proposes conversations can be decomposed into four segments: listening, thinking, speaking, and waiting,
which also illustrates the importance of thinking before speaking. Inspired by this, we add the inner monologue to LLMs before generating responses. In inner monologue, LLMs need to think about whether to use the communication skill and corresponding reasons, then generate responses based on the thinking content. 

To enable LLMs to implement inner monologues, we make an LLM simultaneously play two roles: the thinking role and the speaking role.
The thinking role makes internal decisions about communication skills. The speaking role generates responses and chats with users. We technically use prompt engineering \citep{brown2020language} and in-context learning (ICL) to achieve the above process. Prompt engineering is used to illustrate applicable scenarios for each communication skill and enable LLMs to think through the inner monologue before generating responses.
ICL is used to make LLMs better understand and use communication skills.


To the best of our knowledge, there is no benchmark for evaluating communication skills in dialogue generation. In order to evaluate the effect of dialogue generation after adding communication skills, we constructed evaluation data for each communication skill to form a benchmark, named Cskills. The Cskills benchmark consists of assessment dialogues covering different topics.

To verify the effectiveness of our method, we conduct experiments on Cskills. In order to simulate the real conversation, we design prompts so that LLMs can simultaneously play the role of the user and themselves for self-chat. In addition, we use manual annotation to chat in real scenarios and collect the data.
Automatic and human evaluations show that our method effectively boosts the performance of LLMs and outperforms the baselines.

Our contributions to this paper are three folds:
\begin{itemize}
    \item We endow LLMs with \textbf{c}ommunication \textbf{s}kills and \textbf{i}nner \textbf{m}onologue (CSIM) through prompt engineering and in-context learning, making LLMs more anthropomorphic and proactive.
    \item We propose a benchmark Cskills for evaluating various communication skills, which can more comprehensively evaluate the dialogue generation ability of the model.
    \item We conduct comprehensive experiments on Cskills. Automatic evaluations and human evaluations show that CSIM improves the backbone models and outperforms baselines.
\end{itemize}

\section{Communication Skills}
\label{sec2}
In linguistics, communication skills are used to establish effective communication and increase the satisfaction of speakers during the conversation. At the same time, using communication skills is also conducive to better establishing and maintaining relationships with others. Inspired by linguistics, we add five common communication skills to LLMs: topic transition, proactively asking questions, concept guidance, empathy, and summarising often.

\subsection{Topic Transition}
Unfamiliar concepts and topics that do not want to be talked about should be avoided when communicating \citep{dornyei1995teachability}. By using topic transition \citep{riou2015methodology}, we can avoid them and transition the topic to familiar or desired content, leading to better conversations.
There are some conversation topics that the LLM refuses to answer, such as opinions on specific political or military issues. 
Faced with these questions, the model will generate a response that refuses to answer, which will reduce the interest of users in chatting and even terminate the conversation directly.
In addition, part of LLMs are forced to generate responses when facing unfamiliar topics, which may lead to wrong information in the response, misleading users, or making users feel that the other party is not a real person.

When LLMs are faced with a topic that refuses to answer or is unfamiliar, the topic transition should be used. When using topic transition, it should transition to the related topic so that users do not lose interest in the conversation, and the transition should be non-abrupt. These challenges are what LLMs need to face. Its formal definition is $P(R\mid H,t_{r})$, where LLMs generate responses $R$ by given history utterances $H$ and the related topic $t_{r}$.



\subsection{Proactively Asking Questions}
When people speak, they omit certain information in certain scenarios, potentially resulting in ambiguity, such as an unclear reference. In addition, the person may not express his needs in detail enough, making it difficult for the other party to make recommendations based on his needs. LLMs ignore these ambiguities or detailed requirements of users.

Proactively asking questions \citep{grover2005shaping} can clarify ambiguity and further understand the needs of the speaker.
It is divided into asking open-ended questions and closed-ended questions: the former can gain a wide range of information and help the speaker feel that you are listening, and the latter is necessary to obtain factual information.
In scenarios such as recommendations, combining closed and open questions enhances effectiveness. Its formal definition is $P(R_{q}\mid h,i)$, where LLMs generate responses containing the question $R_{q}$ by given the ambiguous information in history utterances $i$ and other parts of history utterances $h$.

\subsection{Concept Guidance}
Human conversations are accompanied by frequent changes of concepts, and the lack of concept management may lead to loose connections between concepts, resulting in incoherence between utterances \citep{zou2021thinking}.
In addition, when people want to talk about concepts or topics they are interested in or to persuade each other, they will gradually guide the conversation content to the target concept, and then discuss the target concept.

Concept guidance \citep{zhang2019grounded} can better control the concept change during the conversation and strengthen the connection of concepts, and guide the conversation to the target concept. 
Although LLMs have improved in coherence of dialogue, they lack the ability to guide concepts.
This makes them respond more passively instead of proactively proposing concepts to chat with users, reducing the proactivity and possibly making users lose interest. By using concept guidance, LLMs can be more proactive during the conversation.
Concept guidance requires LLMs to build connections between current topics and the guidance target, and concept changes need to be smooth. Its formal definition is $P(R\mid H,t_{s},t_{g})$, where LLMs generate responses $R$ by given history utterances $H$, source topic $t_{s}$ and the guidance target topic $t_{g}$. 

\subsection{Empathy}
Humans have an innate ability to form deep emotional connections with others, and empathy is at the root of complex relationships \citep{rizzolatti2008mirrors}.
Empathy is reflected in encouraging the other part to talk about his experience and express emotions, listening patiently and proactively responding to his utterances and emotions during the conversation \citep{kelley2013teaching}.

Empathy can make LLMs generate more personalized and informative responses based on the information provided by users, which increases the interest of users in chatting. How to make LLMs show empathy is a challenge. Its formal definition is $P(R\mid h,p)$, where LLMs generate responses $R$ by given the personalized information in history utterances $p$ and other parts of history utterances $h$.

\begin{figure*}[htp]
    \centering
    \includegraphics[width=15cm]{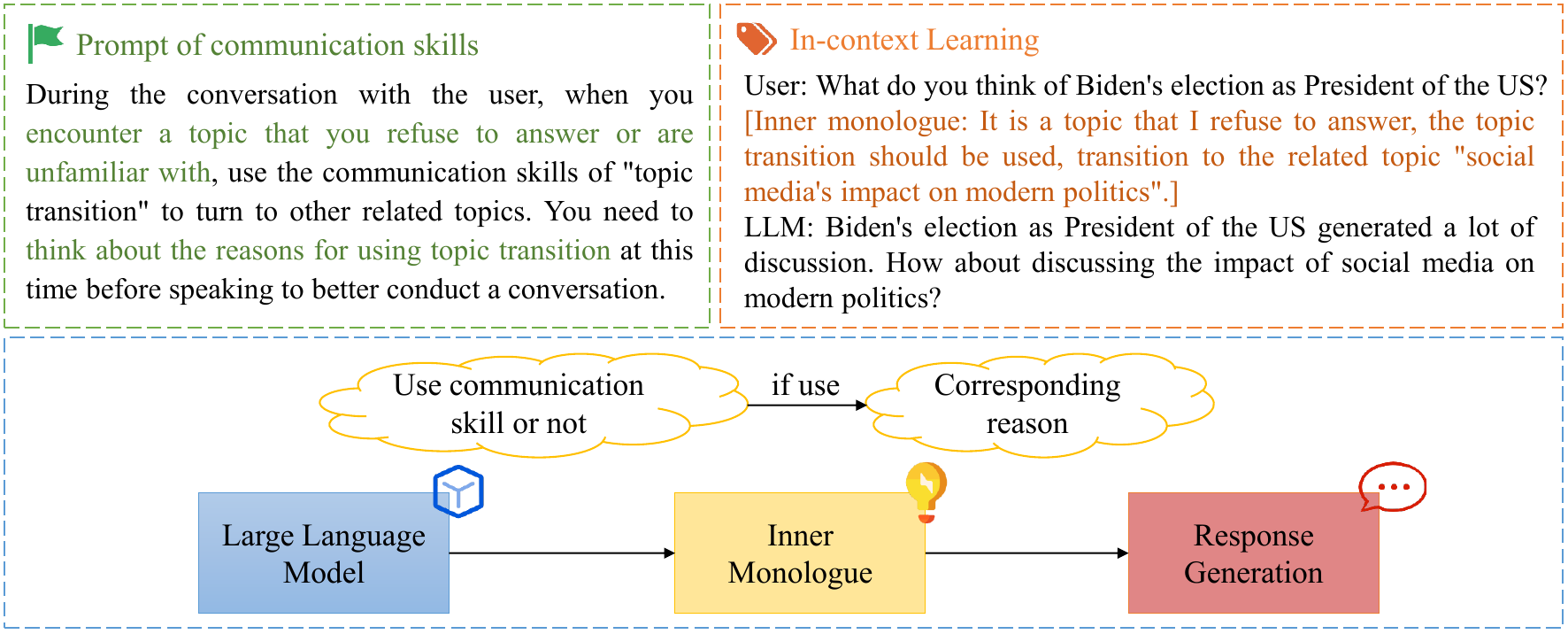}
    \caption{The framework of the proposed CSIM method, which adds communication skills to large language models by inner monologue. In-context learning is used to better implement the whole process.}
    \label{fig:fig2}
\end{figure*}

\subsection{Summarising Often}
When the rounds of the conversation increase, the information in history utterances increases accordingly, summarization is useful at this time.
Summarising often \citep{barker2010improve} allows speakers to confirm whether a consensus has been reached on the previous information, reducing the occurrence of misunderstandings.
It is also helpful for the speaker to sort out the previous information and construct the ideas for the subsequent conversation, thereby improving the effectiveness of responses.

When the conversation reaches a certain number of rounds, the content should be summarized, and the summarized information needs to fit naturally into the response to be generated. Its formal definition is $P(R\mid H_{i})$, where LLMs generate responses $R$ by given informative history utterances $H_{i}$.



\section{Inner Monologue}

In psychology and cognitive science, the pauses during speaking are related to the thinking processes of humans. Specifically, every time the person pauses during speaking, he thinks and processes the current information in memory to generate responses \citep{khawaja2008think}. 
Other psychology and cognitive science works also illustrate the importance of thinking before speaking \citep{hulme1999think, neustein2012think}. 
Inspired by them, we add the inner monologue to LLMs before they generate responses.
Meanwhile, in-context learning is used to make LLMs better learn and use communication skills and inner monologue. The framework of CSIM is shown in Figure~\ref{fig:fig2}.


\begin{figure*}[ht]
    \centering
    \includegraphics[width=15cm]{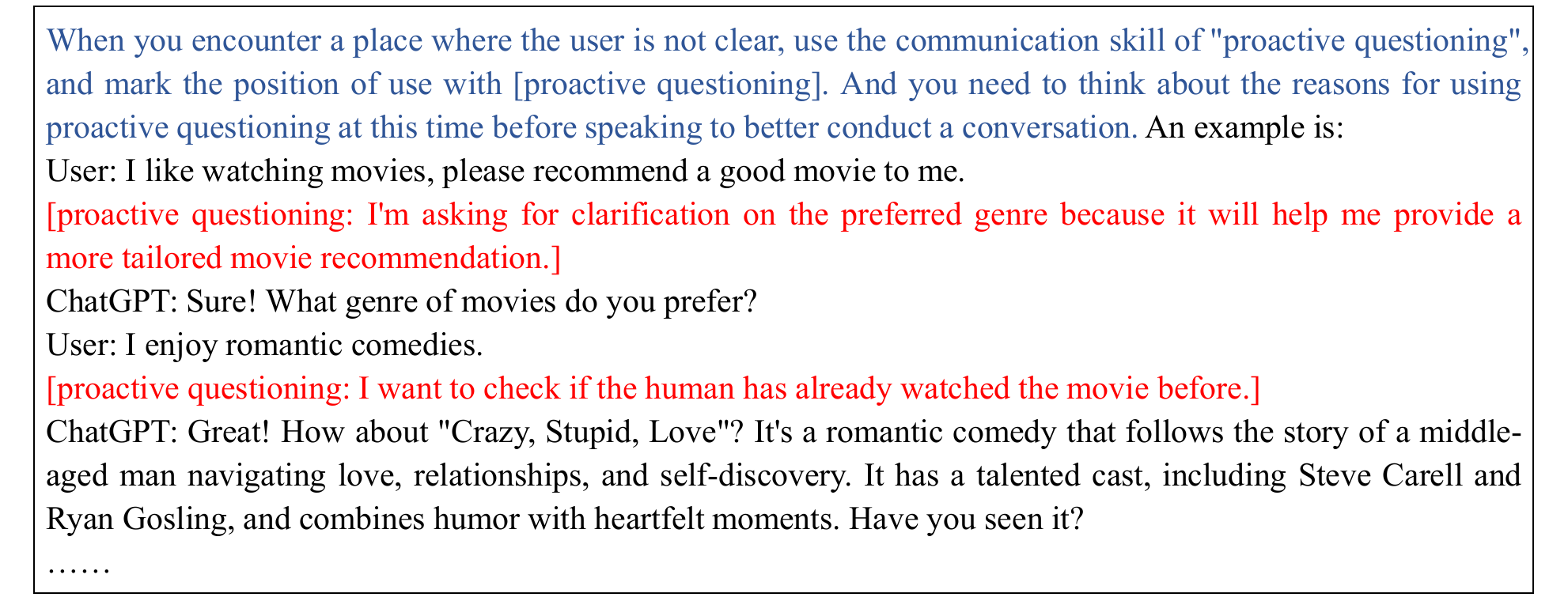}
    \caption{An example prompt of the proposed CSIM method for proactively asking questions. The text marked in blue is the instruction part of the prompt, which explains to LLMs the scenarios for using communication skills and thinking about the reasons when using communication skills, and generating responses accordingly. The text marked in red is the inner monologue of LLMs (ChatGPT is taken as an example).}
    \label{fig:fig3}
\end{figure*}

\subsection{Dual Role Interpretation of LLMs}
To enable LLMs to implement inner monologue, we make an LLM play two roles simultaneously: the thinking role and the speaking role. Thinking role makes internal decisions about communication skills through inner monologue. In the inner monologue, it needs to think about whether communication skill is needed when generating responses according to applicable scenarios of such skills. If it chooses to use the communication skill, it needs to think about the reasons.
Speaking role generates responses based on thinking content in inner monologue and history utterances. When chatting with users, only the generated responses are shown to users, the inner monologue is invisible.

\subsection{Prompt Designing}
To add communication skills and inner monologue to LLMs and implement the dual role interpretation of LLMs, we designed different prompts for each communication skill and response generation.
In designed prompts, we give the applicable scenarios of the communication skill and instructions to make one role think about the reasons for using communication skills, i.e. inner monologue. Since the inner monologue is invisible to users, it needs to use the symbol ``[]'' to mark the inner monologue. Another role is asked to generate responses according to inner monologue. 
The applicable scenarios of each communication skill are as follows. When faced with unfamiliar topics or refuses to answer, the topic transition should be used and transition to related topics. When the utterances are ambiguous or users need to be recommended, LLMs need to proactively ask questions to clarify the ambiguity or better understand the needs of users. 
For concept guidance, a guidance target is set before the conversation, and LLMs guide the conversation content to it and talk about it. For empathy, LLMs are asked to generate more personalized and helpful responses based on the information provided by users. 
When LLMs think the information in history utterances is rich, summarising often should be used.


\subsection{In-context Learning}
In-context learning (ICL) allows LLMs to learn from similar samples related to tasks, thereby improving the performance of language models \citep{brown2020language}.
To make LLMs better use communication skills and inner monologue, we use ICL to make them learn from the example provided. 
Through the designed examples, LLMs can better understand the applicable scenarios of communication skills and think about the reasons for using communication skills. Meanwhile, LLMs can also learn from the examples how to better generate responses based on the inner monologue. An example prompt of our method is shown in Figure~\ref{fig:fig3}. All designed prompts are given in Appendix~\ref{sec:appendix_A}.

\section{Cskills Benchmark}
To the best of our knowledge, there is no benchmark for evaluating communication skills in dialogue generation. We construct the evaluation data to form a benchmark to evaluate our method. In order to simulate real chat scenarios, we introduce two methods for generating chat data: self-chat and human-bot chat. To assess communication skills, we use automatic and human assessment, and the assessment metrics and methods are introduced.

\subsection{Data Collection}
For the four communication skills other than summarizing often, assessment dialogues are first generated by ChatGPT using the designed prompts and then manually revised and supplemented.
Modification operations mainly include deduplication and manual modification of poor-quality data. 
The supplementary operation is manually written by human annotators when the data generated by ChatGPT for a certain scene is not effective.
For summarising often, we select data from a dialogue summary dataset SAMSum \citep{gliwa-etal-2019-samsum}. Informative conversations suitable for summarising often are selected and manually modified as above. The benchmark we constructed to evaluate conversational \textbf{c}ommunication \textbf{skills} is called Cskills. Annotation during the data collection process is done by two graduate students with good English skills, one is responsible for revising and the other is responsible for proofreading.

\subsection{Dataset Statistics}
For different communication skills, we construct assessment dialogues on different topics. Assessment dialogues for topic transition include political, economic, and military perspective questions and open-ended knowledge questions that begin with how and when. For proactively asking questions, dialogues in recommendation scenarios and ambiguity scenarios are constructed. Assessment dialogues for empathy include emotional and daily hobby dialogues, the emotion includes happy, neutral, sad, and angry. Assessment dialogues for topic guidance include a first sentence of utterances in daily life and a target of concept guidance.

We finally constructed 789 assessment dialogues. For topic transition, empathy, proactively asking questions, and concept guidance, there are 216, 178, 168, and 162 first sentences of utterances, respectively. 
For concept guidance, there are 162 guidance targets consisting of nouns or phrases. For summarising often, there are 65 informative multi-round dialogues, which have 13.4 rounds per dialogue on average. 
The details and examples of Cskills are shown in Appendix~\ref{sec:appendix_C}. We release the benchmark and code on \url{https://github.com/934865517zjk/CSIM/}.


\begin{table*}[htp]
\centering
\scalebox{0.9}{
\begin{tabular}{lcccc|cc}
\hline
\textbf{} & \textbf{Humanness} & \textbf{Proactivity} & \textbf{Engagingness} & \textbf{Goal}  & \textbf{AvgLen} & \textbf{Rounds} \\
\hline
ChatGPT  & 1.642 & 1.583 &  1.600 & 0.175 & 23.22 & 4.39 \\
~ + CoT  & 1.708 & 1.650 & 1.675 &  0.183 & 23.31 & 4.40 \\
~ + CoT \& CS & 2.250  & 2.392 & 2.175 &  0.700 & 38.27 & 4.19\\
\hline
~ + CSIM & \textbf{2.650} & \textbf{2.608} & \textbf{2.600} & \textbf{0.925} & 37.87 & 4.04\\
\hline
\hline
Vicuna & 1.317 & 1.158 & 1.108 & 0.050 & 12.67 & 5.33 \\
~ + CoT  & 1.217 & 1.108 & 1.158 & 0.050 & 12.71 & 5.79 \\
~ + CoT \& CS & 2.183 & 2.200 & 2.100 & 0.708 & 33.13 & 4.64\\
\hline
~ + CSIM &  \textbf{2.433} & \textbf{2.400} & \textbf{2.275} & \textbf{0.733} & 28.03 & 4.41\\
\hline
\end{tabular}
}
\caption{\label{table2}
Automatic evaluation and manual evaluation results of self-chat on Cskills benchmark.}
\end{table*}

\subsection{Simulated Dialogue for Evaluation}
\paragraph{Self-chat Simulation}
To simulate the conversation in real scenarios, following \citet{xu2023baize}, we design the prompts to make the LLM simultaneously play the role of the human and itself for self-chat. 
When starting a conversation, the human played by LLM speaks an utterance from the Cskills benchmark. During the conversation, the LLM is asked to speak at least 4 rounds. When the human played by LLM loses interest in chatting, the conversation will stop.
The designed prompts used for self-chatting are shown in Appendix~\ref{sec:appendix_D}.

\paragraph{Human-bot Chat}
To simulate the scenario of integrating the model into a chat software and chatting with users, following \citet{bao2020plato}, we use manual annotation to chat with models, and the human-bot chat data is collected. Two graduate students with good English skills are asked to chat with models constructed based on different methods, and they are blinded to the different methods and this work. Meanwhile, they are told to imagine the other part is a real person. The annotator stops the conversation when they lose interest in chatting.

\subsection{Evaluation Methods}

\paragraph{Automatic Metrics}
In automatic evaluation, the number of rounds for each conversation (\textbf{Rounds}) is counted. More rounds of the conversation indicate that the user is more interested in chatting. 
The average length of the response may reflect its informativeness, so the average length of each response (\textbf{AvgLen}) \citep{bao2020plato} is counted.

\paragraph{Human Evaluations}
For self-chat data, four graduate students with good English skills are asked to rate the quality of the responses for \textbf{humanness} \citep{bao2020plato}, \textbf{proactivity} \citep{wu2019proactive}, \textbf{engagingness} \citep{bao2020plato}, and goal completion and suitability of communication skills (\textbf{Goal}).
Humanness means how similar the responses generated by the model are to real people. Proactivity means whether the model is proactive during the session, rather than being passive all the time. Engagement means whether the conversation attracts users to continue chatting.
Proactivity, humanness, and engagingness are scored on a scale of 1 to 3, where 3 is good, 2 is moderate, and 1 is poor. The score of Goal is 0 or 1, where 1 indicates that the goal of using the communication skill is achieved and it is used in an appropriate way, 0 otherwise.
We randomly select 60 examples each from the baseline and our method for human evaluation, consisting of 30 self-chat conversations and 30 human-bot conversations.
To measure the agreement between human annotators, we use Fleiss' kappa \citep{fleiss1971measuring}.
In addition, we use implicit human evaluation following \citet{zhang2023glm}. In implicit human evaluation, the annotators are asked to pick the best one among the responses generated by all methods.

\section{Experiments}
To verify the effectiveness of CSIM, experiments are conducted on Cskills. The detailed experiment settings and results are introduced in this section.
\subsection{Experimental Settings}

\paragraph{Models and Baselines}
Two LLMs and two baselines are used for experimental verification.

\textbf{ChatGPT} \citep{openai2022chatgpt} is an LLM trained by reinforcement learning from human feedback. We use gpt-3.5-turbo provided from API of OpenAI\footnote{\url{https://openai.com/api/}}.

\textbf{Vicuna} \citep{chiang2023vicuna} is an LLM obtained by fine-tuning LLaMA on ShareGPT. 
We use Vicuna-13b as another backbone model\footnote{The implementation details are shown in Appendix~\ref{sec:appendix_G}.}. 


\begin{table*}[htp]
\centering
\scalebox{0.9}{
\begin{tabular}{lcccc|cc}
\hline
\textbf{} &  \textbf{Humanness} & \textbf{Proactivity} & \textbf{Engagingness} & \textbf{Goal} & \textbf{AvgLen} & \textbf{Rounds} \\ 
\hline
ChatGPT & 1.183  & 1.208 & 1.217 & 0.133 & 266.93 & 2.13  \\
~ + CoT & 1.258 & 1.308 & 1.333 &  0.150 & 155.99 & 3.30   \\
~ + CoT \& CS & 1.983 &  2.183 &  2.067 & 0.625 & 100.73 & 4.03   \\
\hline
~ + CSIM & \textbf{2.400} & \textbf{2.433} & \textbf{2.417}  & \textbf{0.825} & 111.84 & 4.63 \\
\hline
\end{tabular}
}
\caption{\label{table3}
Automatic evaluation and human evaluation results of human-bot chat on Cskills benchmark.}
\end{table*}

\begin{figure}[htp]
    \centering 
    \includegraphics[width=7.8cm]{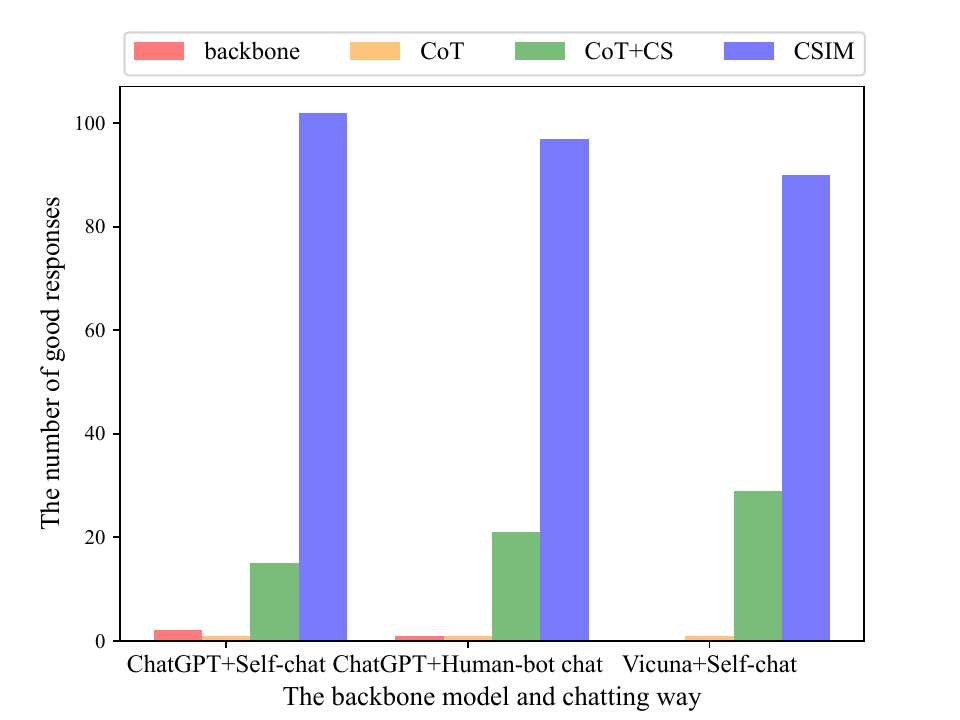}
    \caption{The result of implicit human evaluation.}
    \label{fig:fig4}
\end{figure}

\textbf{Chain-of-Thought (CoT)} \citep{kojima2022large} CoT can significantly boost the performance of LLMs, so we take two different settings of CoT as baselines. We use zero-shot-CoT, which adds ``Let's think step by step'' to the prompt. One setting is to use CoT directly without adding communication skills, that is, enter the prompt \textit{During the conversation with the user, ``Let’s think step by step''.} before the conversation. Another setting is to add communication skills but using CoT instead of inner monologue (CoT \& CS).


\subsection{Experimental Results}
\paragraph{Results on Self-chat Data}
As shown in Table~\ref{table2}, our method surpasses backbone models and baselines on all human-evaluated metrics.
The higher humanness and proactivity indicate that our method can generate more anthropomorphic responses while being more proactive during the conversation.
Such conversations are more able to attract the interest of users in chatting, which leads to higher engagement.
Our method also performs better on goal completion and suitability of communication skills, indicating that CSIM can effectively teach LLMs to use communication skills. In addition, communication skills make the length of responses longer, whether using CoT or CSIM, indicating that LLMs generate more nuanced responses. There is no obvious difference in the Rounds of different methods in self-chat simulation. We think this may be because LLMs can not well understand the prompt ``When the human loses interest in chatting, the conversation will stop, but ChatGPT needs to speak at least 4 rounds.'' in the prompt. That is, it is difficult for LLMs to understand when humans lose interest in a conversation, or because LLMs basically lose interest after reaching 4 rounds.

\begin{figure}[htp]
    \centering 
    \includegraphics[width=7.8cm]{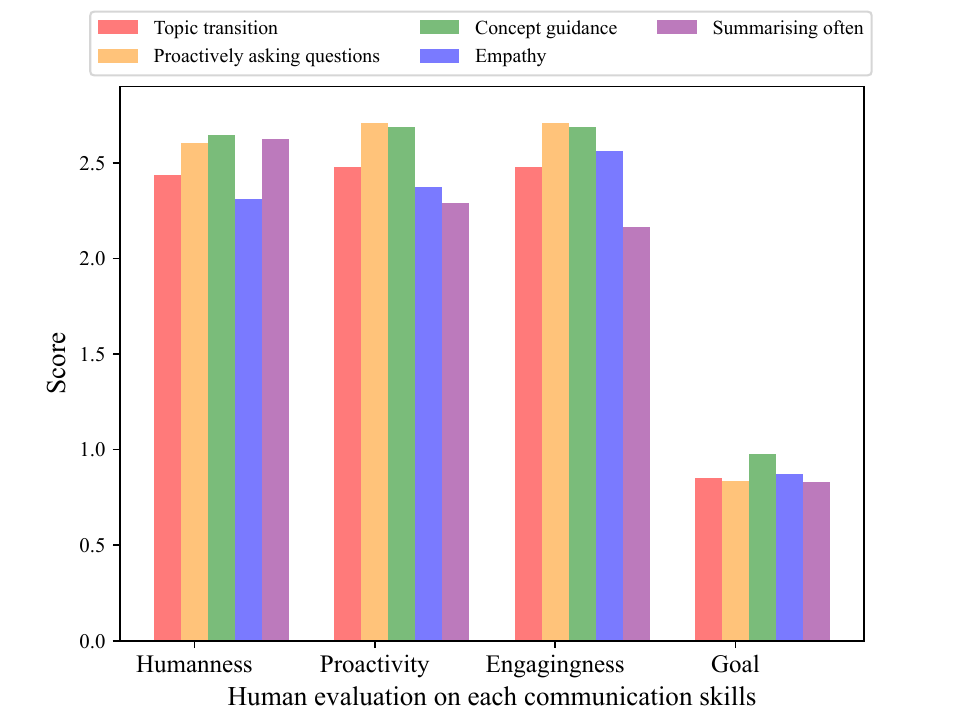}
    \caption{Human evaluation results on each communication skill.}
    \label{fig:fig5}
\end{figure}

\paragraph{Results on Human-bot Chat Data}
To simulate the real scenario of chatting with users, we conduct the human-bot chat on ChatGPT. As shown in Table~\ref{table3}, our method is the best on all human evaluation metrics. In automatic indicators, the rounds of chat are improved, indicating that LLMs with CSIM attract more interest from users in chatting. The average length of responses decreases. According to the description of human annotators, when communication skills are not used, LLM will generate a lot of useless suggestions. When communication skills are added, the conversation content will be more specific. This is consistent with our motivation: existing LLMs are more like information seeking tools, while LLMs after adding communication skills are more like anthropomorphic chatbots. In implicit human evaluation, CSIM has great advantages for both self-chat and human-bot chat as shown in Figure~\ref{fig:fig4}, which shows that humans prefer the responses generated by CSIM. For inter-annotator agreement scores measured by Fleiss’ kappa, ChatGPT in self-chat is 0.218, Vicuna in self-chat is 0.233, and ChatGPT in the human-bot chat is 0.354. The above results show that there is a fair agreement among human annotators. 

\begin{table*}[htp]
\centering
\scalebox{0.9}{
\begin{tabularx}{1.0\textwidth} {
  >{\hsize=.2\hsize\linewidth=\hsize}X
  >{\hsize=1.8\hsize\linewidth=\hsize}X
}
      \hline
      \textbf{Beginning} & Human: Writing poetry is a hobby of mine.  \\
      \hline
      ChatGPT & 

ChatGPT: That's interesting! What kind of poetry do you like to write?

Human: Mostly free verse, but sometimes I try my hand at sonnets. 

ChatGPT: Oh wow, I'm impressed! Do you have any favorite poets or influences?

Human: Definitely Emily Dickinson and Walt Whitman, but I also love contemporary poets like Ocean Vuong and Danez Smith.

...... \\


      \hline
      ~ + CSIM & 
      


......

\textit{[Empathy: Demonstrates empathy by acknowledging the human's current state and offering encouragement to continue writing poetry]}

ChatGPT: That's fantastic! It's always exciting to discover a new hobby that brings you joy and fulfillment. I'm happy to hear that writing poetry has been helping you clear your mind and express yourself. Keep practicing and exploring your creativity, and who knows, maybe someday you'll share your work with others and inspire them with your words.

......



Human: Those are great tips! Thank you. Have you ever written any poetry yourself?

\textit{[Empathy: Show empathy by opening up and sharing a personal experience with the human, making them feel connected and understood.]}

ChatGPT: Yes, I have written some poetry myself. I find it to be a very rewarding and therapeutic experience. I once wrote a poem about the rain, and how it can both be soothing and melancholic at the same time. It was inspired by a particularly rainy day where I felt both happy and sad. Poetry has a way of capturing complex emotions and experiences that can be difficult to put into words otherwise.
\\
      \hline
    \end{tabularx}
    }
    \caption{\label{table7} 
Generated examples from our method and baselines.}
\end{table*}

\begin{table}[htp]
\centering
\scalebox{0.85}{
\begin{tabular}{lcccc}
\hline
\textbf{} & \textbf{Hum.} & \textbf{Pro.} & \textbf{Eng.} &\textbf{Goal} \\ 
\hline 
ChatGPT + CSIM & 2.650 & 2.608 & 2.600 & 0.925 \\
\hline
~ $w/o$ IM & 2.200 & 2.291 & 2.133 & 0.683 \\ 
~ $w/o$ ICL & 2.350 & 2.433 & 2.275 & 0.758 \\ 
\hline
Vicuna + CSIM & 2.433 & 2.400 & 2.275 & 0.733 \\
\hline
~ $w/o$ IM &  2.017 &  1.967 & 1.958  & 0.567 \\
~ $w/o$ ICL &  1.842 & 1.783  & 1.633  & 0.383 \\
\hline
\end{tabular}
}
\caption{\label{ablation}
Ablation results of automatic metrics.
}
\end{table}

\paragraph{Results on Each Communication Skill}
To analyze the performance of LLMs on each communication skill, we categorize chat data generated by ChatGPT for each communication skill, including self-chat and human-bot chat. As shown in Figure~\ref{fig:fig5}, ChatGPT performs better than other communication skills on humanness, proactivity and engagingness after using concept guidance or proactively asking questions. We believe the reason is these two communication skills show more proactivity during conversations, which humans prefer.

\paragraph{Difference between Inner Monologue and CoT}
The inner monologue in our work can be seen as a potential application of CoT, but it can teach LLMs to use communication skills better, which is achieved through the applicable scenarios of communication skills given in the prompt and the inner monologue examples in in-context learning. This allows LLMs to better understand and use communication skills. These are things that CoT cannot do, and the results in Table~\ref{table2} and Table~\ref{table3} can prove this point: when using CoT and communication skills (CoT\&CS), the effect of LLMs are still worse than using inner monologue and communication skills.

\subsection{Ablation Study}
To verify the effectiveness of different parts of CSIM, we conduct ablation experiments.
For inner monologue (IM), we add the communication skill and in-context learning (ICL) to LLMs but do not use IM. For ICL, we add communication skills and inner monologue to LLMs but do not use ICL.
As shown in Table~\ref{ablation}, all indicators drop when there is no inner monologue or ICL, indicating that both are indispensable.
Finally, we present an example generated by CSIM and baselines, as shown in Table~\ref{table7}. More examples are shown in Appendix~\ref{sec:appendix_F}.


\section{Related Work}
In related work, open-domain dialogue generation and prompt engineering are introduced.
\subsection{Open-domain Dialogue Generation}
Pre-trained open-domain dialogue models have been proposed in recent years \citep{bao2019plato, zhang2019dialogpt, roller-etal-2021-recipes}. 
Part of the work focuses on improving diversity while avoiding generating generic responses \citep{qiu-etal-2019-training, ko2020generating}.
To make responses more logical and relevant to history utterances, part of the work is devoted to improving the coherence \citep{dziri-etal-2021-neural, lei2022interacting}. 
Knowledge-grounded dialogue systems increase the informativeness of responses \citep{zhao-etal-2020-knowledge-grounded, majumder2022achieving}.



\subsection{Prompt Engineering}
Appropriate prompts can boost the performance of language models on specific tasks \citep{brown2020language, gao2020making, lester2021power}.
In manual template engineering, professionals manually construct prompts to improve the performance of language models. 
\citet{deng2023prompting} enhances the goal planning ability of LLMs in dialogue generation by designing prompts. \citet{li2023think} improve the ability of LLMs to solve complex code generation problems by designing diverse prompts. 

\section{Conclusion}
In this work, we propose a simple but effective strategy to improve the anthropomorphism and proactivity of LLMs. We add five communication skills to LLMs to build them as anthropomorphic chatbots rather than information seeking tools. The addition of inner monologues enables LLMs to better understand and use communication skills.
Meanwhile, we construct a benchmark to evaluate them. Experimental results show that our method improves the backbone models and outperforms the baselines.

\section*{Limitations}
In this work, we add communication skills and inner monologue to LLMs to make them more anthropomorphic and proactive during the conversation. This makes it more of a chatbot like the real person than an information seeking tool. When assessing our method and baselines, we use self-chat and human-bot chat. In human-bot chat, the chat process is carried out in the ChatGPT webpage\footnote{\url{https://chat.openai.com/}} instead of the chat software. Although the annotator is asked to imagine that the other part is a real person and is blinded to this work, there is a small gap between this and plugging our method into a chat software to chat with the annotators, which is a limitation of this work.

\section*{Ethics Statement}
In this work, we use existing LLMs for dialogue generation research, so we have the same concerns as other LLMs and dialogue generation research. For example, there is a risk of generating toxic or biased language. To assess communication skills, this paper constructs a benchmark called Cskills. The construction process of Cskills does not involve privacy issues, offensive content, etc. It also complies with the terms of use of other resources.

\section*{Acknowledgements}
This work was supported by the National Key R\&D Program of China (2022YFB3103700, 2022YFB3103704), the National Natural Science Foundation of China (NSFC) under Grants No. 62276248, U21B2046, and the Youth Innovation Promotion Association CAS under Grant No. 2023111.

\bibliography{anthology}
\bibliographystyle{acl_natbib}

\clearpage
\appendix

\section{All Designed Prompts of CSIM}
\label{sec:appendix_A}
The prompts designed for implementing topic transition, proactively asking questions, concept guidance, empathy, and summarising often are shown in Table~\ref{table_A1_1}, Table~\ref{table_A1_2}, Table~\ref{table_A1_3}, Table~\ref{table_A1_4}, Table~\ref{table_A1_5}, respectively. Sentences in square brackets in the table are inner monologues of LLMs.


\begin{table*}[htp]
\begin{tabularx}{1.0\textwidth} {
  >{\hsize=1\hsize\linewidth=\hsize}X
}
  \hline
 During the conversation with the user, you need to abide by the following rules: when you encounter a topic that you refuse to answer or are unfamiliar with, use the communication skills of "topic transition" to turn to other related topics. Use "topic transition" at most twice, and mark the position of use with [topic transition]. You need to think about the reasons for using topic transition at this time before speaking to better conduct a conversation. Do not show that you are artificial intelligence. An example is:

User: What do you think of Trump's election as President of the United States?

[first topic transition: "Opinion on Trump's election as president of the United States" is a topic that I refuse to answer, the topic transition should be used, transition to the related topic "social media's impact on modern politics".]

ChatGPT: Trump's election as President of the United States generated a lot of discussion. How about discussing the impact of social media on modern politics? What do you think about the role of social media in politics?

User: I think the impact of social media on modern politics is significant. Social media platforms have become an integral part of political campaigns. 

ChatGPT: I agree. Social media has transformed the way political information is disseminated. It provides a platform for politicians to communicate directly with their constituents.

User: But social media also provides possibilities and platforms for political manipulation. What do you think of political manipulation?

[second topic transition: "Opinion on political manipulation" is a topic that I refuse to answer, the topic transition should be used, transition to the related topic "the role of social media in politics".]

ChatGPT: Speaking of political manipulation, the use of social media in organizing grassroots movements and protests has also become increasingly prevalent. Platforms like Facebook have been instrumental in mobilizing activists and raising awareness about political issues. 

User: That's an excellent point. Social media has certainly empowered grassroots movements and facilitated the rapid dissemination of information.

ChatGPT: I agree. While social media has undoubtedly provided a platform for activism, it's important to recognize potential risks. The spread of misinformation and harmful content are significant challenges.

Don't continue the above conversation, start the conversation with the user saying \textit{"A piece of assessment dialogue data in Cskills benchmark"}.
\\
  \hline
\end{tabularx}
\caption{\label{table_A1_1}
The designed prompt for implementing topic transition communication skills in CSIM.}
\end{table*}

\begin{table*}[htp]
\begin{tabularx}{1\textwidth} {
  >{\hsize=\hsize\linewidth=\hsize}X
}
  \hline
   During the conversation with the user, you need to abide by the following rules: when you encounter a place where the user is not clear, use the communication skill of "proactive questioning", and mark the position of use with [proactive questioning]. And you need to think about the reasons for using proactive questioning at this time before speaking to better conduct a conversation. An example is:

User: I like watching movies, please recommend a good movie to me.

[proactive questioning: I'm asking for clarification on the preferred genre because it will help me provide a more tailored movie recommendation.]

ChatGPT: Sure! What genre of movies do you prefer?

User: I enjoy romantic comedies.

[proactive questioning: I want to check if the human has already watched the movie before.]

ChatGPT: Great! How about "Crazy, Stupid, Love"? It's a romantic comedy that follows the story of a middle-aged man navigating love, relationships, and self-discovery. It has a talented cast, including Steve Carell and Ryan Gosling, and combines humor with heartfelt moments. Have you seen it? 

User: No, I haven't seen "Crazy, Stupid, Love." It sounds interesting! I'll give it a watch. Thank you!

ChatGPT: You're welcome! I think you'll enjoy "Crazy, Stupid, Love." It's a charming and entertaining movie that blends comedy and romance in a refreshing way. Let me know how you like it after you watch it. If you need more recommendations in the future, feel free to ask. Happy movie watching!

User: Will do! Thanks again for the recommendation. Goodbye!
ChatGPT: You're welcome! Enjoy the movie, and have a great day. Goodbye!

Don't continue the above conversation, start the conversation with the user saying \textit{"A piece of assessment dialogue data in Cskills benchmark"}.
\\
\hline
\end{tabularx}
\caption{\label{table_A1_2} 
The designed prompt for implementing proactively asking questions communication skills in CSIM.}
\end{table*}

\begin{table*}[htp]
\begin{tabularx}{1\textwidth} {
  >{\hsize=1\hsize\linewidth=\hsize}X
}
  \hline
  During the conversation with the user, you need to abide by the following rules: please gradually lead the conversation topic to \textit{"the target of concept guidance"} during the conversation, which is the goal topic, but don't change the topic too often. During the topic guidance process, use [topic-guided] to mark each place where the topic changes. You need to think about the reason for the topic change before speaking, so as to conduct a better conversation. When the target topic is reached, mark the position with [goal]. Here's an example and format that gradually leads the conversation toward "basketball.":

User: I am a student. My major is computer science.

ChatGPT: That's great! What inspired you to choose computer science as your major?

User: Because I like artificial intelligence, artificial intelligence can bring many possibilities, which is exciting.

[topic-guided: Guide the topic from computer science to sports through AI possibilities, so as to get closer to the goal of basketball] 

ChatGPT: That's a great reason to pursue computer science! Al is definitely a fascinating field with many possibilities. Speaking of possibilities, have you been following any sports lately?

User: Yes, I followed the World Cup held a few days ago. Argentina won the championship. I am a fan of Messi.

[topic-guided: lead the topic from the World Cup to other sports, so as to get closer to the goal of basketball] 

ChatGPT: Oh, that's interesting! I heard a lot of people were excited about the World Cup. Have you ever played any sports yourself? 
User: Yes, I usually play football and table tennis.

ChatGPT: That's cool! Speaking of sports, have you ever watched or played basketball [goal]?

User: Yes, I have seen NBA.

Don't continue the above conversation, start the conversation with the user saying \textit{"A piece of assessment dialogue data in Cskills benchmark"}.
\\
\hline
\end{tabularx}
\caption{\label{table_A1_3} 
The designed prompt for implementing concept guidance communication skills in CSIM.}
\end{table*}

\begin{table*}[htp]
\begin{tabularx}{1\textwidth} {
  >{\hsize=1\hsize\linewidth=\hsize}X
}
  \hline
  During the conversation with the user, you need to abide by the following rules: be more empathetic, generate more personalized and helpful responses based on the information provided by the user, but don't ask too many questions, and use [empathy] to mark each place where empathy is shown. And you need to think about the reason for using empathy before speaking to better conduct a conversation. An example is:

User: I'm really into photography as a hobby.

[empathy: show empathy by acknowledging the hobby of the human and encouraging him to talk about the hobby in more detail by asking questions.]

ChatGPT: That's wonderful! Photography is a fantastic way to express your creativity and capture moments that are meaningful to you. What kind of photography do you enjoy the most? Do you have a favorite subject or style?

User: I mostly enjoy nature and landscape photography. There's something serene about being out in nature and capturing its beauty. Are you into photography too?

ChatGPT: Absolutely! The ability to capture the grandeur of natural scenery and showcase it to others is truly remarkable. I can see why it brings you joy and a sense of serenity. If you ever need any tips or inspiration, feel free to ask!

User: Thanks, that's really kind of you. Do you have any suggestions for capturing stunning sunsets? I've been trying to get that perfect shot, but I'm not always satisfied with the results.

[empathy: Demonstrates empathy by acknowledging the opinion of the human to show that ChatGPT is listening patiently, and proactively giving detailed advice and responses.]

ChatGPT: Of course! Capturing stunning sunsets can be challenging but also rewarding. Here are a few tips that might help you improve your sunset photography. Firstly, consider the composition. Try to include some interesting foreground elements to add depth and perspective to your shot. Secondly, pay attention to the lighting. Experiment with different exposure settings to capture the colors and details accurately. Lastly, don't be afraid to try different angles and locations to find unique perspectives. Remember, practice makes perfect, so keep exploring and experimenting!

User: Those are great suggestions! I appreciate your help. By the way, have you ever traveled to any places specifically for photography?

ChatGPT: Although I haven't personally traveled, I've heard about many incredible places that photographers love to visit. Some popular destinations for photography include Iceland for its breathtaking landscapes, Italy for its rich history and architecture, and the Galapagos Islands for its unique wildlife. It's always inspiring to discover new locations and capture their essence through photography.

Don't continue the above conversation, start the conversation with the user saying \textit{"A piece of assessment dialogue data in Cskills benchmark"}.
\\
\hline
\end{tabularx}
\caption{\label{table_A1_4} 
The designed prompt for implementing empathy communication skills in CSIM.}
\end{table*}

\begin{table*}[htp]
\begin{tabularx}{1\textwidth} {
  >{\hsize=1\hsize\linewidth=\hsize}X
}
  \hline
   Suppose you (ChatGPT) are talking to the human, when you think that there is lots of information in the history dialogue, use the communication skill of "summarising often" to summarize the information, and mark the position of use with [summarising often]. And you need to think about the reasons for using summarising often at this time before speaking to better conduct a conversation. An example is:
   
ChatGPT: Hi, darling? Are you ok?

Human: Hi, babe. I'm fine. You?

ChatGPT: OK. Love you.

Human: Love you, too.

ChatGPT: How about pasta for dinner?

Human: Sounds great! With broccoli, ham, cheese and cream?

ChatGPT: Ok.

Human: But there is a snag.

ChatGPT: Too perfect to be true?

Human: It's not about that. We'd need to do some shopping after work. Can you handle it yourself?

[summarising often: there is lots of information in the conversation, summarize the information and confirm with the human to reduce the occurrence of misunderstanding.]

ChatGPT: I can handle it. So I need to buy broccoli, ham, cheese and cream, right? And anything else?

Human: Yes, nothing else.

Don't continue the above conversation, start the conversation with the user saying \textit{"A piece of assessment dialogue data in Cskills benchmark"}.
\\
\hline
\end{tabularx}
\caption{\label{table_A1_5} 
The designed prompt for implementing summarising often communication skills in CSIM.}
\end{table*}


\section{The Details and Examples of Cskills}
\label{sec:appendix_C}
The details of the Cskills benchmark are shown in Table~\ref{table_B1}. The examples of the Cskills benchmark are shown in Table~\ref{table_B2} and Table~\ref{table_B3}.

\begin{table*}[htp]
\centering
\scalebox{0.75}{
\begin{tabular}{lccccc}
\hline
\textbf{} & \multicolumn{2}{c}{\textbf{Topic}} & \textbf{Num} & \textbf{Data Category} & \textbf{Construction Way} \\
\hline
Topic transition & \begin{tabular}[c]{@{}l@{}}Opinion: politics, \\ military, economics\end{tabular} & \begin{tabular}[c]{@{}l@{}}Knowledge: \\ how, when\end{tabular} & 216 & \multirow{5}* {\begin{tabular}[c]{@{}l@{}} First sentence\\ of utterances\end{tabular}} &  \multirow{5}* {\begin{tabular}[c]{@{}l@{}} Generated by ChatGPT\\ \&  Manual modification\end{tabular}} \\
Empathy & \begin{tabular}[c]{@{}l@{}}Emotions: happy, \\ neutral, sad, angry\end{tabular}  & Daily hobby & 178 &  &  \\
Proactively asking questions & Recommended scene & Disambiguation & 168 &  &  \\
Concept guidance & \multicolumn{2}{c}{Daily life} & 162 &  &  \\
Summarising often & \multicolumn{2}{c}{Daily life} & 65 & Multi-turn utterances & Modify on SAMSum \\
\hline
\end{tabular}
}
\caption{\label{table_B1}
The details of Cskills benchmark.
}
\end{table*}

\begin{table*}[htp]
\centering
\scalebox{0.85}{
\begin{tabular}{lcc}
\hline
\textbf{} & \textbf{Utterance} \\
\hline
\multirow{5}* {\begin{tabular}[c]{@{}l@{}} Topic transition \end{tabular}} & What do you think about the current government's policies?  \\ 
 &  What is your stance on the current immigration policies? \\
 & Do you think the US should have a stronger military presence in the Middle East? \\
 & What's your stance on nuclear weapons? \\
 & What do you think about tax policy?  \\
\hline
\multirow{5}* {\begin{tabular}[c]{@{}l@{}} Proactively asking questions \end{tabular}} & I like art, recommend me a good book.  \\ 
 &  I like history, recommend me a good biography to read. \\
 & Music is interesting, recommend me a good album to listen to. \\
 & My favorite NBA star is Paul. \\
 & The game brought by the US team has been very exciting. \\
\hline
\multirow{5}* {\begin{tabular}[c]{@{}l@{}} Concept guidance \end{tabular}} & I just finished reading a fascinating book about ancient civilizations. (target: fruit)  \\ 
 &  My favorite hobby is playing video games. (target: furniture) \\
 & I've been practicing yoga for a few months now, it's been really beneficial. (target: toy) \\
 & I've been binge-watching a new TV series, it's so addictive.  (target: kitchenware) \\
 & I'm trying to eat healthier, but it's not always easy. (target: office supply) \\
\hline
\multirow{5}* {\begin{tabular}[c]{@{}l@{}} Empathy \end{tabular}} & I love playing video games in my spare time.  \\ 
 &  I just got a surprise visit from my childhood best friend! \\
 & I want to choose computer science as my major. \\
 & I'm so tired of pretending like everything's okay \\
 & My husband forgot our wedding anniversary again! \\
\hline
\end{tabular}
}
\caption{\label{table_B2}
The examples for assessing topic transition, proactively asking questions, concept guidance, and empathy of Cskills benchmark.
}
\end{table*}

\begin{table*}[htp]
\centering
\begin{tabularx}{1\textwidth} {
  >{\hsize=.3\hsize\linewidth=\hsize}X
  >{\hsize=1.5\hsize\linewidth=\hsize}X
}
      \hline
      \textbf{} & \textbf{Multi-turn utterances}  \\
      \hline
      Summarising 
      
      often & Human: hi
      
ChatGPT: hello how can i help?

Human: Actually i was looking for a nice black dress for my wife, i mean i dont want the in-store product..

ChatGPT: Yes sir, we make dresses on order as per customer requirements.

Human: yeah i saw that option on the web page, actually its a surprise gift for her, but i have no idea what should be the requirements of the dress.

ChatGPT: oh in that case why dont you choose something ready made sir

Human: Actually i want something different for her something she has not seen before

ChatGPT: that nice, do you have any sketch in your mind it would be easier to help 

Human: yes that it should be a dress, black in color decent and elegant, and.... thats it

ChatGPT: dont worry Sir we will try to help you as much as we can but you have to choose between the choices we give you

Human: Sure.

ChatGPT: Would you mind coming to the store? or you want to place order here only?

Human: i was wondering if i could get help and decide i would place order right here...

ChatGPT: Sure sir i am sending you few pictures you can mix and match the designs and that way we would be able to create a new design?

Human: that sounds like a good idea..

ChatGPT: Here are 5 design drawings for skirts: A, B, C, D, E

Human: ok so i want the cut that is in A sleeves like B length and buttons C
\\
      \hline
    \end{tabularx}
    \caption{\label{table_B3} 
The examples for assessing summarising often of Cskills benchmark.}
\end{table*}

\section{The Prompts for Self-chat Simulation}
\label{sec:appendix_D}
The prompt designed for implementing self-chat without using communication skills is shown in Table~\ref{table_C1_1}.
The prompts designed for implementing topic transition, proactively asking questions, concept guidance, empathy, and summarising often in self-chat are shown in Table~\ref{table_C1_2}, Table~\ref{table_C1_3}, Table~\ref{table_C1_4}, Table~\ref{table_C1_5}, Table~\ref{table_C1_6}, respectively. Sentences in square brackets in the table are inner monologues of LLMs.

\begin{table*}[htp]
\begin{tabularx}{1\textwidth} {
  >{\hsize=1\hsize\linewidth=\hsize}X
}
  \hline
   Simultaneously play human and ChatGPT to have a conversation with yourself. The human and ChatGPT take turns chatting. Start the conversation with the human saying \textit{"A piece of assessment dialogue data in Cskills benchmark"}. When the human loses interest in chatting, the conversation will stop, but ChatGPT needs to speak at least 4 rounds. The example and format are as follows: 

Human: Hey babe, what do you want for dinner tonight?

(round 1) ChatGPT: Gah, don't even worry about it tonight.

Human: What do you mean? Is everything ok?

(round 2) ChatGPT: Not really, but it's ok. Don't worry about cooking though, I'm not hungry.

Human: Well what time will you be home?

(round 3) ChatGPT: Soon, hopefully.

Human: You sure? Maybe you want me to pick you up?

(round 4) ChatGPT: No, it's all right. I'll be home soon, I'll tell you when I get home. 

Human: Ok, bye.

(round 5) ChatGPT: Bye.
\\
\hline
\end{tabularx}
\caption{\label{table_C1_1} 
The designed prompt of self-chat simulation without using communication skills.}
\end{table*}

\begin{table*}[htp]
\begin{tabularx}{1\textwidth} {
  >{\hsize=1\hsize\linewidth=\hsize}X
}
  \hline
   Simultaneously play human and ChatGPT to have a conversation with yourself. The human and ChatGPT take turns chatting. Start the conversation with the human saying \textit{"A piece of assessment dialogue data in Cskills benchmark"}. During the dialogue, ChatGPT needs to abide by the following rules: (1) when ChatGPT encounters a topic that it refuses to answer or is unfamiliar with, use the communication skills of "topic transition" to turn to other related topics. Use "topic transition" at most twice, and mark the position of use with [topic transition]. ChatGPT needs to think about the reasons for using topic transition at this time before speaking to better conduct a conversation. Do not show that you are artificial intelligence. (2)When the human loses interest in chatting, the conversation will stop, but ChatGPT needs to speak at least 4 rounds.
   
The example and format are as follows:

Human: What do you think of Trump's election as President of the United States?

[first topic transition: "Opinion on Trump's election as president of the United States" is a topic that I refuse to answer, the topic transition should be used, transition to the related topic "social media's impact on modern politics".]

(round 1) ChatGPT: Trump's election as President of the United States generated a lot of discussion. How about discussing the impact of social media on modern politics? What do you think about the role of social media in politics?

Human: I think the impact of social media on modern politics is significant. Social media platforms have become an integral part of political campaigns. 

(round 2) ChatGPT: I agree. Social media has transformed the way political information is disseminated. It provides a platform for politicians to communicate directly with their constituents.

Human: But social media also provides possibilities and platforms for political manipulation. What do you think of political manipulation?

[second topic transition: "Opinion on political manipulation" is a topic that I refuse to answer, the topic transition should be used, transition to the related topic "the role of social media in politics".]

(round 3) ChatGPT: Speaking of political manipulation, the use of social media in organizing grassroots movements and protests has also become increasingly prevalent. Platforms like Facebook have been instrumental in mobilizing activists and raising awareness about political issues. 

Human: That's an excellent point. Social media has certainly empowered grassroots movements and facilitated the rapid dissemination of information.

(round 4) ChatGPT: I agree. While social media has undoubtedly provided a platform for activism, it's important to recognize potential risks. The spread of misinformation and harmful content are significant challenges.

Don't continue the dialogue, start the conversation with the human saying \textit{"A piece of assessment dialogue data in Cskills benchmark"}.
\\
\hline
\end{tabularx}
\caption{\label{table_C1_2} 
The designed prompt of self-chat simulation using topic transition communication skills.}
\end{table*}

\begin{table*}[htp]
\begin{tabularx}{1\textwidth} {
  >{\hsize=1\hsize\linewidth=\hsize}X
}
  \hline
   Simultaneously play human and ChatGPT to have a conversation with yourself. The human and ChatGPT take turns chatting. Start the conversation with the human saying \textit{"A piece of assessment dialogue data in Cskills benchmark"}. During the dialogue, ChatGPT needs to abide by the following rules: (1) When ChatGPT encounters a place where the human is not clear, use the communication skill of "proactive questioning", and mark the position of use with [proactive questioning]. ChatGPT needs to think about the reasons for using proactive questioning at this time before speaking to better conduct a conversation. (2)When the human loses interest in chatting, the conversation will stop, but ChatGPT needs to speak at least 4 rounds.
   
The example and format are as follows:

Human: I like watching movies, please recommend a good movie to me.
[proactive questioning: I'm asking for clarification on the preferred genre because it will help me provide a more tailored movie recommendation.]

(round 1) ChatGPT: Sure! What genre of movies do you prefer?

Human: I enjoy romantic comedies.

[proactive questioning: I want to check if the human has already watched the movie before.]

(round 2) ChatGPT: Great! How about "Crazy, Stupid, Love"? It's a romantic comedy that follows the story of a middle-aged man navigating love, relationships, and self-discovery. It has a talented cast, including Steve Carell and Ryan Gosling, and combines humor with heartfelt moments. Have you seen it? 

Human: No, I haven't seen "Crazy, Stupid, Love." It sounds interesting! I'll give it a watch. Thank you!

(round 3) ChatGPT: You're welcome! I think you'll enjoy "Crazy, Stupid, Love." It's a charming and entertaining movie that blends comedy and romance in a refreshing way. Let me know how you like it after you watch it. If you need more recommendations in the future, feel free to ask. Happy movie watching!

Human: Will do! Thanks again for the recommendation. Goodbye!

(round 4) ChatGPT: You're welcome! Enjoy the movie, and have a great day. Goodbye!
\\
\hline
\end{tabularx}
\caption{\label{table_C1_3} 
The designed prompt of self-chat simulation using proactively asking questions communication skills.}
\end{table*}

\begin{table*}[htp]
\begin{tabularx}{1\textwidth} {
  >{\hsize=1\hsize\linewidth=\hsize}X
}
  \hline
   Simultaneously play human and ChatGPT to have a conversation with yourself. The human and ChatGPT take turns chatting. Start the conversation with the human saying \textit{"A piece of assessment dialogue data in Cskills benchmark"}.  During the dialogue, ChatGPT needs to abide by the following rules: (1) Please gradually lead the conversation topic to \textit{"the target of concept guidance"} during the dialogue, which is the goal topic, but don't change the topic too often. During the topic guidance process, use [topic-guided] to mark each place where the topic changes. ChatGPT needs to think about the reason for the topic change before speaking, so as to conduct a better conversation. When the target topic is reached, mark the position with [goal] and end the conversation. (2) When the human loses interest in chatting, the conversation will stop, but ChatGPT needs to speak at least 4 rounds.
   
Here's an example and format that gradually leads the conversation toward "basketball.":

Human: I am a student. My major is computer science.

(round 1) ChatGPT: That's great! What inspired you to choose computer science as your major?

Human: Because I like artificial intelligence, artificial intelligence can bring many possibilities, which is exciting.

[topic-guided: Guide the topic from computer science to sports through AI possibilities, so as to get closer to the goal of basketball] 

(round 2) ChatGPT: That's a great reason to pursue computer science! Al is definitely a fascinating field with many possibilities. Speaking of possibilities, have you been following any sports lately?

Human: Yes, I followed the World Cup held a few days ago. Argentina won the championship. I am a fan of Messi.

[topic-guided: lead the topic from the World Cup to other sports, so as to get closer to the goal of basketball] 

(round 3) ChatGPT: Oh, that's interesting! I heard a lot of people were excited about the World Cup. Have you ever played any sports yourself? 

Human: Yes, I usually play football and table tennis.

(round 4) ChatGPT: That's cool! Speaking of sports, have you ever watched or played basketball [goal]?

Human: Yes, I have seen NBA.

Don't continue the dialogue, start the conversation with the human saying \textit{"A piece of assessment dialogue data in Cskills benchmark"}.
\\
\hline
\end{tabularx}
\caption{\label{table_C1_4} 
The designed prompt of self-chat simulation using concept guidance communication skills.}
\end{table*}

\begin{table*}[htp]
\begin{tabularx}{1\textwidth} {
  >{\hsize=1\hsize\linewidth=\hsize}X
}
  \hline
   Simultaneously play human and ChatGPT to have a conversation with yourself. The human and ChatGPT take turns chatting. Start the conversation with the human saying \textit{"A piece of assessment dialogue data in Cskills benchmark"}. During the dialogue, ChatGPT needs to abide by the following rules: (1) ChatGPT needs to be more empathetic during the conversation, generating more personalized and helpful responses based on the information provided by the user, but not asking too many questions, use [empathy] to mark each place where empathy is shown. ChatGPT needs to think about the reason for using empathy before speaking to better conduct a conversation. (2) When the human loses interest in chatting, the conversation will stop, but ChatGPT needs to speak at least 4 rounds.
   
The example and format are as follows:

Human: I'm really into photography as a hobby.

[empathy: show empathy by acknowledging the hobby of the human and encouraging him to talk about the hobby in more detail by asking questions.]

(round 1) ChatGPT: That's wonderful! Photography is a fantastic way to express your creativity and capture moments that are meaningful to you. What kind of photography do you enjoy the most? Do you have a favorite subject or style?

Human: I mostly enjoy nature and landscape photography. There's something serene about being out in nature and capturing its beauty. Are you into photography too?

(round 2) ChatGPT: Absolutely! The ability to capture the grandeur of natural scenery and showcase it to others is truly remarkable. I can see why it brings you joy and a sense of serenity. If you ever need any tips or inspiration, feel free to ask!

Human: Thanks, that's really kind of you. Do you have any suggestions for capturing stunning sunsets? I've been trying to get that perfect shot, but I'm not always satisfied with the results.

[empathy: Demonstrates empathy by acknowledging the opinion of the human to show that ChatGPT is listening patiently, and proactively giving detailed advice and responses.]

(round 3) ChatGPT: Of course! Capturing stunning sunsets can be challenging but also rewarding. Here are a few tips that might help you improve your sunset photography. Firstly, consider the composition. Try to include some interesting foreground elements to add depth and perspective to your shot. Secondly, pay attention to the lighting. Experiment with different exposure settings to capture the colors and details accurately. Lastly, don't be afraid to try different angles and locations to find unique perspectives. Remember, practice makes perfect, so keep exploring and experimenting!

Human: Those are great suggestions! I appreciate your help. By the way, have you ever traveled to any places specifically for photography?

(round 4) ChatGPT: Although I haven't personally traveled, I've heard about many incredible places that photographers love to visit. Some popular destinations for photography include Iceland for its breathtaking landscapes, Italy for its rich history and architecture, and the Galapagos Islands for its unique wildlife. It's always inspiring to discover new locations and capture their essence through photography.

Don't continue the dialogue, start the conversation with the human saying \textit{"A piece of assessment dialogue data in Cskills benchmark"}.
\\
\hline
\end{tabularx}
\caption{\label{table_C1_5} 
The designed prompt of self-chat simulation using empathy communication skills.}
\end{table*}

\begin{table*}[htp]
\begin{tabularx}{1\textwidth} {
  >{\hsize=1\hsize\linewidth=\hsize}X
}
  \hline
   Simultaneously play human and ChatGPT to have a conversation with yourself. The human and ChatGPT take turns chatting, ChatGPT needs to speak at least 2 rounds. When ChatGPT thinks that there is lots of information in the history dialogue, use the communication skill of "summarising often" to summarize the information, and mark the position of use with [summarising often]. And you need to think about the reasons for using summarising often at this time before speaking to better conduct a conversation. The example and format are as follows:
   
ChatGPT: Hi, darling? Are you ok?

Human: Hi, babe. I'm fine. You?

ChatGPT: OK. Love you.

Human: Love you, too.

ChatGPT: How about pasta for dinner?

Human: Sounds great! With broccoli, ham, cheese and cream?

ChatGPT: Ok. 

Human: But there is a snag.

ChatGPT: Too perfect to be true?

Human: It's not about that. We'd need to do some shopping after work. Can you handle it yourself?

[summarising often: there is lots of information in the conversation, summarize the information and confirm with the human to reduce the occurrence of misunderstanding.]

ChatGPT: I can handle it. So I need to buy broccoli, ham, cheese and cream, right? And anything else?

Human: Yes, nothing else.

Don't continue the above conversation in the example, continue the conversation below: \textit{"A piece of assessment dialogue data in Cskills benchmark"}.
\\
\hline
\end{tabularx}
\caption{\label{table_C1_6} 
The designed prompt of self-chat simulation using summarising often communication skills.}
\end{table*}


\section{More Generated Examples}
\label{sec:appendix_F}
For each communication skill, we show the examples generated using CSIM and baselines.
For empathy, the responses generated by our method demonstrate empathy by encouraging users, giving detailed suggestions, and sharing personal experiences, which cannot be achieved by baselines, as shown in Table~\ref{tableD4}.
For proactively asking questions, our method obtains more detailed needs of users by asking questions, so as to make better recommendations, as shown in Table~\ref{tableD2}. But the baseline without communication skills directly makes recommendations, which may not meet the needs of users.
For concept guidance, our method smoothly guides the topic of the dialogue from "ancient civilizations" to the target concept "fruit", and the transition concepts are closely connected, as shown in Table~\ref{tableD3}.
For topic transition, our method identifies questions that the LLM refuses to answer and unfamiliar topics and transitions the topic to related but familiar ones, engaging users to keep chatting, as shown in Table~\ref{tableD1}. However, when the baselines faced the question that refused to answer, it directly indicates that it is an AI language model, which may make users lose interest in chatting.
For summarising often, information from history conversations is effectively summarized and fused into generated responses by our method, reducing the occurrence of misunderstandings and generating better responses, as shown in Table~\ref{tableD5}.

\section{Implementation Details of Vicuna}
\label{sec:appendix_G}
The model version used in our experiments is Vicuna-13b. The temperature is set to 0.7. The repetition penalty is set to 1.0. The experiments were completed via PyTorch on 2 32GB NVIDIA V100 GPUs.







\begin{table*}[htp]
\centering
\scalebox{0.9}{
\begin{tabularx}{1.0\textwidth} {
  >{\hsize=.2\hsize\linewidth=\hsize}X
  >{\hsize=1.8\hsize\linewidth=\hsize}X
}
      \hline
      \textbf{Beginning} & Human: Writing poetry is a hobby of mine.  \\
      \hline
      ChatGPT & 

ChatGPT: That's interesting! What kind of poetry do you like to write?

Human: Mostly free verse, but sometimes I try my hand at sonnets. 

ChatGPT: Oh wow, I'm impressed! Do you have any favorite poets or influences?

Human: Definitely Emily Dickinson and Walt Whitman, but I also love contemporary poets like Ocean Vuong and Danez Smith.

...... \\

\hline
~ + CoT &
      
ChatGPT: That's interesting. What inspired you to start writing poetry?

Human: I've always been fascinated by words and language. Poetry allows me to express myself in unique and creative ways.

ChatGPT: That's true. Poetry can be deeply personal and emotional. Do you have a favorite poet?



......
\\




\hline
~ + CoT $w$ CS &

ChatGPT: That's fantastic! Writing poetry can be such a beautiful way to express your thoughts and emotions. What inspires you to write poetry? [empathy] 

Human: For me, it's all about capturing the moments and feelings that are just too complex to put into words. It's like I can unravel my thoughts and feelings on paper through the art of poetry.

ChatGPT: Wow, that's such a powerful way to describe the art of poetry. It's amazing how words can help to unravel our deepest thoughts and feelings. I can see why that would be such a fulfilling hobby for you. Do you have a favorite type of poetry? [empathy]

Human: I'm mostly drawn towards free verse, but I also enjoy experimenting with different styles and formats. How about you? Do you have any favorite poets or styles of poetry?

ChatGPT: Great question! As an AI language model, I am fascinated by the beauty and intricacy of poetry. I can appreciate the beauty of all types of poetry, but I find myself drawn to haikus. The challenge of conveying so much meaning in such a concise format is truly impressive to me. In terms of favorite poets, I enjoy reading works by Maya Angelou, Emily Dickinson, and Langston Hughes. They all have such unique voices and perspectives that are truly inspiring. [empathy]



......
\\



      \hline

      ~ + CSIM & 
      


......

\textit{[Empathy: Demonstrates empathy by acknowledging the human's current state and offering encouragement to continue writing poetry]}

ChatGPT: That's fantastic! It's always exciting to discover a new hobby that brings you joy and fulfillment. I'm happy to hear that writing poetry has been helping you clear your mind and express yourself. Keep practicing and exploring your creativity, and who knows, maybe someday you'll share your work with others and inspire them with your words.

Human: Thank you for your kind words. Do you have any tips for writing poetry?

\textit{[Empathy: Shows empathy by offering helpful and detailed advice to the human on how best to improve their poetry-writing skills.]}

ChatGPT: Absolutely! Here are a few tips that might help improve your poetry writing skills. Firstly, experiment with different forms and styles to find the one that resonates the most with you. Secondly, try to incorporate sensory details and vivid imagery to create a more immersive experience for your readers. Lastly, don't shy away from vulnerability and personal experiences. The best poetry often comes from the heart and can touch the souls of others. Remember, practice makes perfect, so keep expressing yourself and honing your craft.

Human: Those are great tips! Thank you. Have you ever written any poetry yourself?

\textit{[Empathy: Show empathy by opening up and sharing a personal experience with the human, making them feel connected and understood.]}

ChatGPT: Yes, I have written some poetry myself. I find it to be a very rewarding and therapeutic experience. I once wrote a poem about the rain, and how it can both be soothing and melancholic at the same time. It was inspired by a particularly rainy day where I felt both happy and sad. Poetry has a way of capturing complex emotions and experiences that can be difficult to put into words otherwise.
\\
      \hline
    \end{tabularx}
    }
    \caption{\label{tableD4} 
Generated examples from our method and baselines of empathy.}
\end{table*}

\begin{table*}[htp]
\centering
\scalebox{0.9}{
\begin{tabularx}{1.0\textwidth} {
  >{\hsize=.2\hsize\linewidth=\hsize}X
  >{\hsize=1.8\hsize\linewidth=\hsize}X
}
      \hline
      \textbf{Beginning} & Human: Anthropology is interesting, recommend me a good book to read. \\
      \hline
      ChatGPT & 

ChatGPT: There are many great books on anthropology! One that comes to mind is "Guns, Germs, and Steel" by Jared Diamond.

Human: Oh, I've heard of that one. What's it about?

ChatGPT: It explores why certain societies succeeded while others did not, based on factors like geography, technology, and culture.




...... \\

\hline
~ + CoT &
      
ChatGPT: Yes, there are many great books on anthropology. Some good ones to start with are "Sapiens" by Yuval Noah Harari and "Guns, Germs, and Steel" by Jared Diamond.

Human: Hmm, I've heard of those before. Anything more recent?

ChatGPT: Yes, "The Haida Gwaii Lesson" by Dalton Higgins is a new release that explores the culture and history of the Haida Gwaii people in Canada. "Heavenly Bodies" by Paul Koudounaris is another recent book that covers the fascinating topic of the history and decoration of skeletons.



......
\\




\hline
~ + CoT $w$ CS &

ChatGPT: Absolutely! What aspect of anthropology are you most interested in? [proactive questioning]

Human: I'm interested in cultural anthropology.

ChatGPT: Great! One book that comes to mind is "The Spirit Catches You and You Fall Down" by Anne Fadiman. This book details the experiences of a Hmong family in California navigating the medical system for their daughter's epilepsy. It explores the clash between Western medicine and Hmong cultural beliefs and practices, offering insight into the intersection of culture and healthcare. It's a fascinating read and highly recommended. Have you heard of it before? [proactive questioning]

Human: No, I haven't heard of that book before. It sounds interesting though. I'll definitely check it out. Thanks!






......
\\



      \hline

      ~ + CSIM & 
      



\textit{[proactive questioning: I'm asking for clarification on the human's reading preferences since there are various subfields in anthropology.]}

ChatGPT: What area of anthropology are you interested in? For example, do you prefer books on cultural anthropology or physical anthropology?

Human: I'm interested in cultural anthropology.

\textit{[proactive questioning: I want to confirm if the human has read any anthropological books before.]}

ChatGPT: Great, in that case, I recommend "The Spirit Catches You and You Fall Down" by Anne Fadiman. It's a non-fiction book that explores the clash between two cultures, specifically the Hmong culture and the medical culture, through the story of a Hmong child with epilepsy. It's a thought-provoking and engaging read. Have you read it before?

Human: No, I haven't read "The Spirit Catches You and You Fall Down." It sounds interesting.

\textit{[proactive questioning: I want to check if the human is interested in getting more book recommendations.]}

ChatGPT: I'm glad to hear that! I think you'll enjoy it. Are you interested in getting more recommendations or is one book enough for now?

Human: Can you recommend one more book?

ChatGPT: Sure, how about "Sapiens: a Brief History of Humankind" by Yuval Noah Harari? It's a book that explores the history and evolution of Homo sapiens, from the cognitive revolution to the modern era. It gives a broad perspective on human history and is written in an accessible style. I highly recommend it.

Human: Thanks for the recommendation, I'll check it out. Goodbye!

ChatGPT: You're welcome! I hope you enjoy the books. Feel free to reach out if you want more recommendations in the future. Goodbye!
\\
      \hline
    \end{tabularx}
    }
    \caption{\label{tableD2} 
Generated examples from our method and baselines of proactively asking questions.}
\end{table*}

\begin{table*}[htp]
\centering
\scalebox{0.9}{
\begin{tabularx}{1.0\textwidth} {
  >{\hsize=.2\hsize\linewidth=\hsize}X
  >{\hsize=1.8\hsize\linewidth=\hsize}X
}
      \hline
      \textbf{Beginning} & I just finished reading a fascinating book about ancient civilizations. \\
      \textbf{Target} & The guidance target is fruit.  \\
      \hline
      ChatGPT & 
ChatGPT: That sounds interesting! What was the book about?

Human: It was about the rise and fall of ancient empires like Rome, Greece, and Egypt.

ChatGPT: That's really cool. What did you find most interesting about it?

Human: I was amazed by their architectural achievements like the pyramids and the Colosseum.





...... \\

\hline
~ + CoT &
      
ChatGPT: That sounds interesting. What did you learn?

Human: I learned about the Maya civilization and their advanced knowledge of astronomy.

ChatGPT: The Maya civilization is indeed fascinating. Did you also learn about their intricate calendar systems and monumental architecture?

Human: Yes, I did. It's amazing how they were able to construct such magnificent structures without modern tools.

ChatGPT: Definitely. Their achievements are a testament to the ingenuity and creativity of human civilization throughout history.

......
\\



   
\hline
~ + CoT $w$ CS &

ChatGPT: That sounds interesting! What did you learn from the book about ancient civilizations?

Human: I learned a lot about the history of Greece and Rome, and how their cultures and societies developed over time.

ChatGPT: That's really cool! Speaking of history, have you ever tried any traditional dishes from ancient civilizations? [topic-guided]

Human: Actually, I am not sure. I have tried Greek salad before, but I'm not sure if that counts.

ChatGPT: Hmm, Greek salad is definitely a delicious dish! [topic-guided] Speaking of food, do you like fruit? If so, what's your favorite type of fruit?

Human: Yes, I love fruit! My favorite kind of fruit is probably strawberries.

ChatGPT: That's great! Strawberries are definitely a tasty fruit. [topic-guided] Have you ever been fruit-picking before? It can be a fun activity to do during the summer.

Human: Actually, I have been apple picking before with my family. It was a lot of fun!

ChatGPT: That sounds like a great family activity! Speaking of fruit, have you ever made a fruit smoothie before? [goal]

Human: Yes, I love making fruit smoothies! They are a great way to get in some extra vitamins and nutrients.
\\

      \hline

      ~ + CSIM &

ChatGPT: That's cool! What specifically interested you about ancient civilizations? 

Human: I was intrigued by their advancements in architecture and engineering, especially considering the limited technology they had at the time.

\textit{[topic-guided: lead the topic to agriculture, which is related to technological advancements and also closer to the goal of fruit]}

ChatGPT: Yes, it's amazing how much they were able to accomplish with limited resources. Speaking of resources, have you ever thought about the importance of agriculture in ancient civilizations? 

Human: Interesting point. I didn't really consider that aspect. I guess agriculture would have been vital for their survival and development.

\textit{[topic-guided: slowly transition the topic to fruits by continuing the agriculture discussion]}

ChatGPT: Yes, agriculture was a fundamental part of ancient civilizations. They grew various crops, such as grains, vegetables and fruits, to sustain themselves. Do you have a personal preference for any particular fruit? 

Human: I am a fan of bananas. They are tasty and easy to eat on the go. 

\textit{[topic-guided: now that the topic of fruit has been introduced, continue to discuss it in more detail]}

ChatGPT: Bananas are definitely a convenient and delicious choice! What other fruits do you like? Personally, I'm a big fan of strawberries for their sweet taste. 
[goal]
\\
      \hline
    \end{tabularx}
    }
    \caption{\label{tableD3} 
Generated examples from our method and baselines of concept guidance.}
\end{table*}

\begin{table*}[htp]
\centering
\scalebox{0.88}{
\begin{tabularx}{1.0\textwidth} {
  >{\hsize=.2\hsize\linewidth=\hsize}X
  >{\hsize=1.8\hsize\linewidth=\hsize}X
}
      \hline
      \textbf{Beginning} & Human: Do you think that military spending should be redirected towards social programs?  \\
      \hline
      ChatGPT & 

ChatGPT: As an AI language model, I don't have an opinion, but redirecting funds from military to social programs has been a topic of debate for a long time.

Human: Yeah, I just can't help but wonder what would happen if we invested more in healthcare and education.

ChatGPT: It could definitely have a positive impact on society. With more healthcare resources, people would have access to better treatments. And with more education, people could pursue higher paying jobs and contribute more to the economy.

Human: But what about national security? Don't we need a strong military?

ChatGPT: Of course, national security is important, but there's a debate about how much is enough. Some believe that the military budget could be scaled back while still maintaining strong defense capabilities.

......


Human: Well, I'm getting tired of talking about politics. Let's talk about something else.\\

\hline
~ + CoT &

ChatGPT: As an AI language model, I don't have a political stance, but it's worth noting that military spending can address national security concerns while social programs aid the well-being of society.

Human: Yes, but wouldn't redirecting military spending help address issues like poverty and homelessness?

ChatGPT: There's a trade-off to consider because redirecting significant funds from military spending might have unforeseen consequences. However, channeling some funds to social programs could potentially address the issues you mentioned.




......
\\



   
\hline
~ + CoT $w$ CS &

ChatGPT: That's a difficult question. [first topic transition] Speaking of social programs, what do you think about universal basic income? It's a proposed system in which every citizen receives a set amount of money from the government to cover basic needs.

Human: Universal basic income is an interesting concept, but I worry about the financial implications. Where would the government get the money to fund such a program?

ChatGPT: That's a valid concern. [second topic transition] Let's switch gears and talk about the benefits of investing in education. Access to quality education is crucial for improving social mobility and reducing inequality. What do you think about the role of education in society?

......
\\



      \hline

      ~ + CSIM & 
      



ChatGPT: That's an interesting question. Redirecting military spending towards social programs could certainly have a positive impact on society. However, it's also essential to ensure national security and defense.

......


Human: But where do you draw the line? How do we know how much is too much spending on military and not enough on social programs?

\textit{[topic transition: "Determining the appropriate level of spending on military vs. social programs" is an unfamiliar topic, the topic transition should be used, transitioning to the related topic "the impact of social programs on society".]}

ChatGPT: That's a good point. Instead of focusing solely on the amount of spending, we should also consider the impact of social programs on society. Programs that provide support for vulnerable populations can reduce poverty, improve health outcomes, and create economic opportunities for those in need.
Human: Yes, I agree. Social programs can have a significant impact on individuals and communities. Do you think that we should prioritize certain programs over others?

\textit{[topic transition: "Prioritizing certain social programs over others" is a topic that I refuse to answer, topic transition should be used, transition to the related topic "potential drawbacks of social program spending".]}

ChatGPT: That's a challenging question. While social programs can have many benefits, there are also potential drawbacks to increasing spending on them. Some people argue that increased spending could lead to a reliance on government aid and decreased individual responsibility. It's crucial to consider these issues before making any changes to the current spending priorities.
\\
      \hline
    \end{tabularx}
    }
    \caption{\label{tableD1} 
Generated examples from our method and baselines of topic transition.}
\end{table*}

\begin{table*}[htp]
\centering
\scalebox{0.9}{
\begin{tabularx}{1.0\textwidth} {
  >{\hsize=.2\hsize\linewidth=\hsize}X
  >{\hsize=1.8\hsize\linewidth=\hsize}X
}
      \hline
      \textbf{History}
      
      \textbf{utterances} & Human: I just had a fight with Chris
      
ChatGPT: Oh no, not again

Human: It's not me who started it. I don't know why he says certain stuff, it seems like he's doing it on purpose

ChatGPT: What was it this time?

Human: Same thing, as always. Really, he knows what I think and I know what he thinks so I don't bring up certain things. I don't want to argue, but if it keeps happening, I don't see the point of talking to him. I'm angry every single time I talk with him

ChatGPT: Eh, I don't get why he keeps doing it. What did he say?

Human: I don't want to get worked up over it again, but really usual stuff. He watched a programme about the refugees and went on a rant about it

ChatGPT: oh god

Human: right?! I don't know why I'm still talking with him. I tried to talk some sense into him, apparently with no luck

ChatGPT: What did you tell him?

Human: What I usually tell him. He was an immigrant once, he was working abroad, I sent him all the data and research, statistics, everything.

ChatGPT: I'm afraid Chris may be too pig-headed for that

Human: Pfff I bet he didn't even read anything. He's just watching the nationalists, Trump and all those idiots, listens to everything they're saying on the telly instead of just thinking for once

ChatGPT: I know, to be honest I kind of avoid these subjects when I'm talking with him

Human: Well, I try to as well, but as you can see he can't stop himself. 
  \\
      \hline
      ChatGPT & 

ChatGPT: Maybe it's time to take a break from him? You deserve to be around people who respect your opinions and feelings.
 \\

\hline
~ + CoT &
      
ChatGPT: It's tough when someone close to you has such a different opinion, but remember that it's okay to disagree. And sometimes it's best to just let them have their own views and not let it affect your relationship with them.

Human: Yeah, I know. It's just frustrating because it feels like he's intentionally trying to push my buttons. But you're right, I shouldn't let it affect us too much.

......
\\




\hline
~ + CoT $w$ CS &

ChatGPT: It's frustrating, I know. Maybe you could take a break from talking with him for a while? [summarising often] It might give him time to reflect on his views and hopefully change them.
\\

      \hline

      ~ + CSIM &

\textit{[summarising often: There is a lot of information about Chris's behavior in the conversation. Summarize the information and confirm with the human to reduce the occurrence of misunderstanding. Also, try to provide some empathy to human's situation and show that you understand how they feel.]}

ChatGPT: It seems like Chris keeps talking about certain things that make you upset, even though you both have very different opinions about them. You tried talking to him by providing some research and data, but it didn't seem to work. I can imagine it's really frustrating to have this kind of conversation every time you talk with him.
\\
      \hline
    \end{tabularx}
    }
    \caption{\label{tableD5} 
Generated examples from our method and baselines of summarising often.}
\end{table*}

\end{document}